\def\year{2023}\relax

\documentclass[letterpaper]{article} 
\usepackage{aaai23}  
\usepackage{times}  
\usepackage{helvet}  
\usepackage{courier}  
\usepackage[hyphens]{url}  
\usepackage{graphicx} 
\usepackage{natbib}  
\usepackage{caption} 
\usepackage{subcaption}
\usepackage{booktabs}
\usepackage{multirow}
\DeclareCaptionStyle{ruled}{labelfont=normalfont,labelsep=colon,strut=off} 
\usepackage[inline]{enumitem}
\usepackage{siunitx}
\usepackage{amsmath}
\usepackage{amssymb}
\usepackage[section]{placeins}

\title{Out-Of-Distribution Detection Is Not All You Need}

\author{
Joris Guérin, \textsuperscript{\rm 1, 2, 4} 
Kevin Delmas, \textsuperscript{\rm 3} 
Raul Sena Ferreira, \textsuperscript{\rm 1, 2} 
Jérémie Guiochet \textsuperscript{\rm 1, 2}.
}

\affiliations{
\textsuperscript{\rm 1}Université de Toulouse,
\textsuperscript{\rm 2}LAAS-CNRS,
\textsuperscript{\rm 3}ONERA, Toulouse, France \\
\textsuperscript{\rm 4}Espace-Dev, IRD, univ. Montpellier, Montpellier, France \\
joris.guerin@ird.fr, kevin.delmas@onera.fr, rsenaferre@laas.fr, jeremie.guiochet@laas.fr
}

\begin{document}

\maketitle


\begin{abstract}
The usage of deep neural networks in safety-critical systems is limited by our ability to guarantee their correct behavior. Runtime monitors are components aiming to identify unsafe predictions and discard them before they can lead to catastrophic consequences. Several recent works on runtime monitoring have focused on out-of-distribution (OOD) detection, i.e., identifying inputs that are different from the training data. In this work, we argue that OOD detection is not a well-suited framework to design efficient runtime monitors and that it is more relevant to evaluate monitors based on their ability to discard incorrect predictions. We call this setting out-of-model-scope detection and discuss the conceptual differences with OOD. We also conduct extensive experiments on popular datasets from the literature to show that studying monitors in the OOD setting can be misleading:
\begin{enumerate*}
    \item very good OOD results can give a false impression of safety,
    \item comparison under the OOD setting does not allow identifying the best monitor to detect errors.
\end{enumerate*}  
Finally, we also show that removing erroneous training data samples helps to train better monitors.
\end{abstract}


\section{Introduction} \label{sec:intro}

\begin{figure*}[t]
    \centering
    \begin{subfigure}[b]{0.48\textwidth}
        \centering
        \includegraphics[scale=0.18]{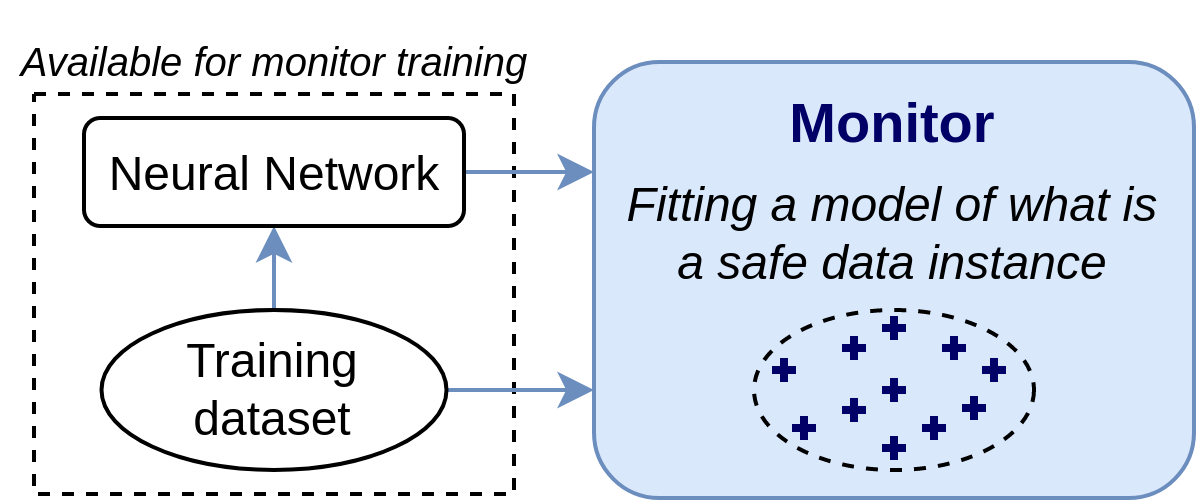}
        \caption{Monitor training (design time)}
        \label{fig:RM_def_training}
    \end{subfigure}
~
    \begin{subfigure}[b]{0.48\textwidth}
        \centering
        \includegraphics[scale=0.18]{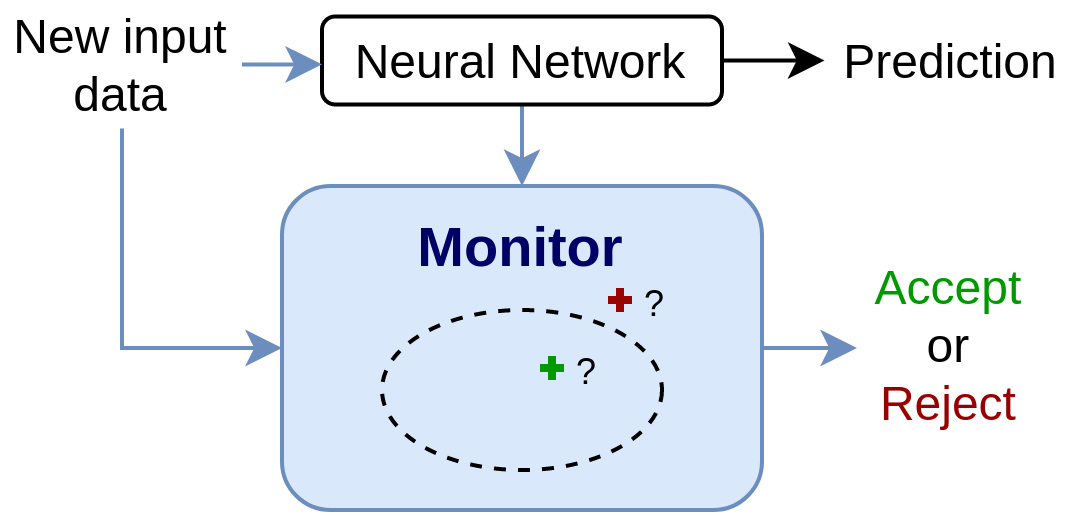}
        \caption{Using monitor at inference time (runtime)}
        \label{fig:RM_def_inference}
    \end{subfigure}
    \caption{\textbf{Neural Network Runtime Monitoring} is a research area aiming to characterize safe data instances, in order to reject unsafe predictions during inference. Different definitions of \emph{safe data instances} gave rise to different subfields in the literature.}
    \label{fig:RM_def}
\end{figure*}

With the recent progress in machine learning (ML) research, deep neural network (DNN) architectures are now used to address safety-critical tasks, 
e.g., self-driving cars~\cite{stocco2020misbehaviour}, surgical robots~\cite{surgical_robotics}, drones landing~\cite{guerin2022evaluation}. 
Online fault tolerance approaches, or runtime monitors, are promising research directions to improve the safety of such systems. 
A runtime monitor is a component aiming to identify and reject unsafe data encountered at inference time. As in most of the recent literature, this paper focuses on the unsupervised setting, where we do not have access to examples of ``unsafe input data'' during monitor training. In other words, to separate safe data instances from unsafe ones, we need to fit a one-class classifier~\cite{khan2014occ_review} using only the DNN training dataset. Figure~\ref{fig:RM_def} illustrates the life cycle of a DNN runtime monitor (training and inference phases).

Many approaches have been proposed to tackle the DNN runtime monitoring problem defined above. However, in the literature, they are found under different names as they adopt different definitions of ``unsafe data instances''. On the one hand, the field of Out-Of-Distribution (OOD) detection aims at identifying input data that are not from the training distribution~\cite{wang2022vim, sun2021react, schorn2020facer}. 
On the other, several works consider directly the problem of identifying input data that lead to errors of the monitored DNN~\cite{granese2021doctor, wang2020dissector, kang2018model}. In this work, we name this second view Out-of-Model-Scope (OMS) detection. In practice, the approaches to address OOD and OMS detection are not different and follow the same workflow: they use the DNN training dataset to build a one-class classifier (the monitor) to characterize safe data instances and use it to reject unsafe samples (Figure~\ref{fig:RM_def}). These two paradigms only differ in their objectives (definition of normal data samples), and by extension in how new approaches are evaluated. We emphasize that OOD and OMS are evaluation settings and not monitoring approaches per se.

In this work, we claim that the real goal of a DNN runtime monitor is to tackle the OMS problem, i.e., to identify prediction errors before they can propagate through the system. We argue that OOD detection was designed as a proxy task for OMS detection, based on the belief that what the DNN knows is equivalent to the information contained in the dataset that was used to train it. The first objective of this paper is to discuss the conceptual differences between the OOD and OMS detection settings. Then, we conduct experiments to determine whether OOD detection is a good proxy for OMS detection. In other words, we want to know whether the OMS performance of a runtime monitor can be correctly assessed under the OOD setting. Finally, based on the insights gained from these discussions, we formalize a good practice to train better monitors. We show experimentally that, when fitting the monitor, it is better to remove erroneous samples from the DNN training dataset.

This paper is organized as follows. In section~\ref{sec:definitions}, notations are introduced and the concepts of OMS and OOD are clearly defined. In section~\ref{sec:related}, we present the existing literature about OMS and OOD detection. In section~\ref{sec:differences_theory}, we discuss the differences between OOD and OMS and explain why the OOD setting can lead to an inaccurate perceived performance of a monitor. In section~\ref{sec:differences_expe}, we conduct experiments to verify whether these limitations of OOD detection actually occur in practical scenarios. Finally, in section~\ref{sec:recommendations}, we introduce a new training trick to build better monitors.

\section{Notations and definitions}\label{sec:definitions}
Let $T$ be an ML task, defined by an oracle function $\Omega$, on a domain $\mathcal{X}$, i.e., $\forall x \in \mathcal{X}$, the ground-truth for $T$ is $\Omega(x)$. This work mostly discusses classification, but could easily be extended to other ML tasks. Let $f$ be a DNN used to solve $T$, and let $m_f$ be a monitor of $f$, i.e., a one-class classifier used during inference to reject unsafe predictions of $f$.

\subsection{The Out-of-Model-Scope detection setting}\label{sec:oms_definition}
We define the scope of $f$, $\mathcal{D}_f$, to be the set of data instances where $f$ is correct:
\begin{equation}\label{eq:model_scope}
    D_f = \{x \in \mathcal{X} \, \vert \, f(x) = \Omega(x)\}
\end{equation}
Ideally, we want $m_f$ to identify data points that lead to errors of $f$. The perfect monitor for $f$, noted $m_f^*$, is defined by
\begin{equation}\label{eq:optimal_monitor_oms}
    \forall x \in \mathcal{X}, \; m_f^*(x) = 
    \begin{cases} 
    0 & \text{if } x \in \mathcal{D}_f, \\
    1 & \text{else}.
\end{cases}
\end{equation}
We call Out-of-Model-Scope (OMS) detection, the task of designing a monitor that reproduces the behavior of $m_f^*$. OMS detection is defined with respect to a specific model $f$. Indeed, if another neural network $f'$ (different architecture or weights) is used, the model scope will likely be different ($\mathcal{D}_{f'} \neq \mathcal{D}_{f}$), and so will the optimal monitor ($m_{f'}^* \neq m_f^*$).

\subsection{The Out-Of-Distribution detection setting}\label{sec:ood_definition}
In practice, the monitored DNN $f$ is trained using a supervised training dataset, i.e., a small subset of $n$ data points in $\mathcal{X}$ for which the ground truth is known:
\begin{equation*}
    D_{\text{train}} = \{(x_i, y_i) \; | \; \forall i \in \{1, ...n\} \, x_i \in \mathcal{X}, y_i = \Omega(x_i)\}.
\end{equation*}
A common practice for DNN monitoring is to define an in-distribution domain $\mathcal{D}_{\text{ID}}$, that comprises all data instances drawn from the same distribution as $D_{\text{train}}$. The Out-Of-Distribution (OOD) detection task aims to build a monitor $m$ to identify data instances that do not belong to $\mathcal{D}_{\text{ID}}$, i.e., the perfect OOD monitor ($m^*$) is defined by
\begin{equation}\label{eq:optimal_monitor_ood}
    \forall x \in \mathcal{X}, \; m^*(x) = 
    \begin{cases} 
    0 & \text{if } x \in \mathcal{D}_{\text{ID}}, \\
    1 & \text{else}.
\end{cases}
\end{equation}
The rationale behind OOD detection is that DNNs trained on $D_{\text{train}}$ must be good for input data similar to $D_{\text{train}}$ ($x \in \mathcal{D}_{\text{ID}}$), but should not perform well on other data ($x \notin \mathcal{D}_{\text{ID}}$). The evaluations conducted in the OOD setting are independent of $f$, they only depend on the task and training dataset.


\section{Related works} \label{sec:related}

This section presents several monitoring approaches. Conceptually, OOD and OMS detectors are not different, i.e., one-class classifiers fitted to $D_{\text{train}}$, usually using feature representations extracted from $f$ (Figure~\ref{fig:RM_def}). However, in practice, they differ in the way they are evaluated experimentally. 

\subsection{OMS in the literature} \label{sec:related_oms}
Several previous works have studied the problem of OMS detection, i.e., they developed DNN runtime monitoring approaches and assessed their performance based on their ability to detect errors of $f$. 
\citet{granese2021doctor} developed 
a monitor called DOCTOR, which computes an optimal rejection score based on the softmax vector returned by $f$.
\citet{wang2020dissector} proposed a monitor called DISSECTOR, which rejects inputs with incoherent activations among different layers of $f$.
The monitor proposed by \citet{kang2018model} is specific to object detection. They developed a model assertion technique to specify constraints on the shape of the predicted bounding boxes.
\citet{cheng2019runtime} proposed to store neuron activations of $D_{\text{train}}$ 
in the form of binary decision diagrams, and reject patterns that were not seen previously.


\subsection{OOD detection in the literature} \label{sec:related_ood}
The definition of OOD in equation~\ref{eq:optimal_monitor_ood} is frequently seen in the literature. However, the boundaries of $\mathcal{D}_{\text{ID}}$ are fuzzy and there is no clear definition of whether a data point was drawn from the same distribution as $D_{\text{train}}$. 
To overcome this issue, OOD detection works consider that the test split associated with $D_{\text{train}}$ belongs to $\mathcal{D}_{\text{ID}}$. Then, different approaches exist to construct OOD sets, and the monitors are evaluated based on their ability to distinguish whether a data point is from the test set or the OOD set. In the literature, three main concepts of ``OODness'' are used to build OOD datasets.

\paragraph{Novelty} A data point $x \in \mathcal{X}$ is OOD if the ground-truth $\Omega(x)$ is not among the predefined classes handled by $f$, i.e., $f$ cannot be correct (e.g., an image of a frog presented to a cat vs. dog classifier). A large body of works was developed and evaluated in this configuration. 
\citet{liang2018odin} proposed ODIN, an approach 
using input preprocessing and temperature scaling on the softmax outputs to maximize ID/OOD separation. 
\citet{liu2020energy} proposed an energy score ($logsumexp$ of the logits) to detect OOD data. 
\citet{sun2021react} showed that clipping the highest values of the last layer can help to build better logits for OOD detection. Their approach is called ReAct and can be combined with any logits-based scores.
\citet{lee2018mahalanobis} proposed to fit Gaussian distributions to features extracted from $f$, and 
to use the Mahalanobis distance as an OOD score.
All the above approaches used another dataset, with disjoint label classes, as the OOD set.
Another idea to build OOD sets is to remove data samples from certain classes while training $f$ and $m_f$, and use them as OOD examples to test the monitor. 
This was done by \citet{henzinger2020outside}, who developed a monitor called Outside-the-Box, using the smallest bounding boxes containing all features representing $D_{\text{train}}$ to model the boundary of $\mathcal{D}_{\text{ID}}$.

\paragraph{Covariate shift} OOD data instances present different characteristics than $D_{\text{train}}$, but with valid ground-truth. 
Covariate shift comes from changes in external conditions, sensor failures, or modifications to the environment. Such threats to DNN safety 
were discussed extensively by~\citet{hendrycks2019benchmarking}. Most works dealing with covariate shift detection have built OOD sets by injecting perturbations to test images. 
\citet{schorn2020facer} 
injected different kinds of noise patterns and evaluated their monitor called FACER. 
\citet{cai2020real} 
used a variational auto-encoder and anomaly detection to detect abnormal rain conditions (injected in simulation). 
\citet{chen2020task} 
applied brightness changes to the test images to build an OOD set to test their approach. 
Finally, \citet{zhang2018deeproad} used distances in the feature space of a 
DNN 
to separate sunny images (used for training) from rainy or snowy images (from YouTube).

\paragraph{Adversarial attack} A data sample that was modified to make a DNN fail with high confidence. The difference with covariate shift is in the malicious intent to generate imperceptible perturbations. Several works presented above also tested their approaches for adversarial attack detection. 
In addition, \citet{kantaros2021real} proposed to detect adversarial attacks by testing the robustness of a prediction against input image transformations.
Similarly, \citet{wang2019adversarial} considered the robustness 
against random mutations of $f$.

\subsection{Approaches considering both OMS and OOD} \label{sec:related_both}
\citet{hendrycks17baseline} 
proposed to use the Maximum Softmax Probability (MSP) as a rejection score. They evaluated the performance of MSP under both the OOD and OMS settings.
\citet{ferreira2021benchmarking} conducted a benchmark study of OOD detection methods. Their experiments included novelties, covariate shifts, and adversarial attacks. They also reported the OMS detection performance. However, in both of these papers, OOD and OMS are simply viewed as separate problems. Here, we consider that OOD is simply a proxy for OMS, and we discuss whether both paradigms are useful for the field of DNN monitoring.


\section{Limitations of OOD detection} \label{sec:differences_theory}
In this work, we claim that a successful runtime monitor $m_f$ should perform well under the OMS setting, i.e., reject data samples that correspond to errors of $f$, and accept others. DNNs are usually good on their training dataset (train, validation, and test splits), while their performance decreases when data move away from $D_{\text{train}}$~\cite{hendrycks2019benchmarking}. These two facts gave rise to the highly studied problem of OOD detection, aiming to reject samples that are far from the training distribution. Studying the alternate OOD detection problem in place of OMS detection presents several conceptual issues, which are highlighted in this section.

\begin{figure}[t]
    \centering
    \includegraphics[width=0.48\textwidth]{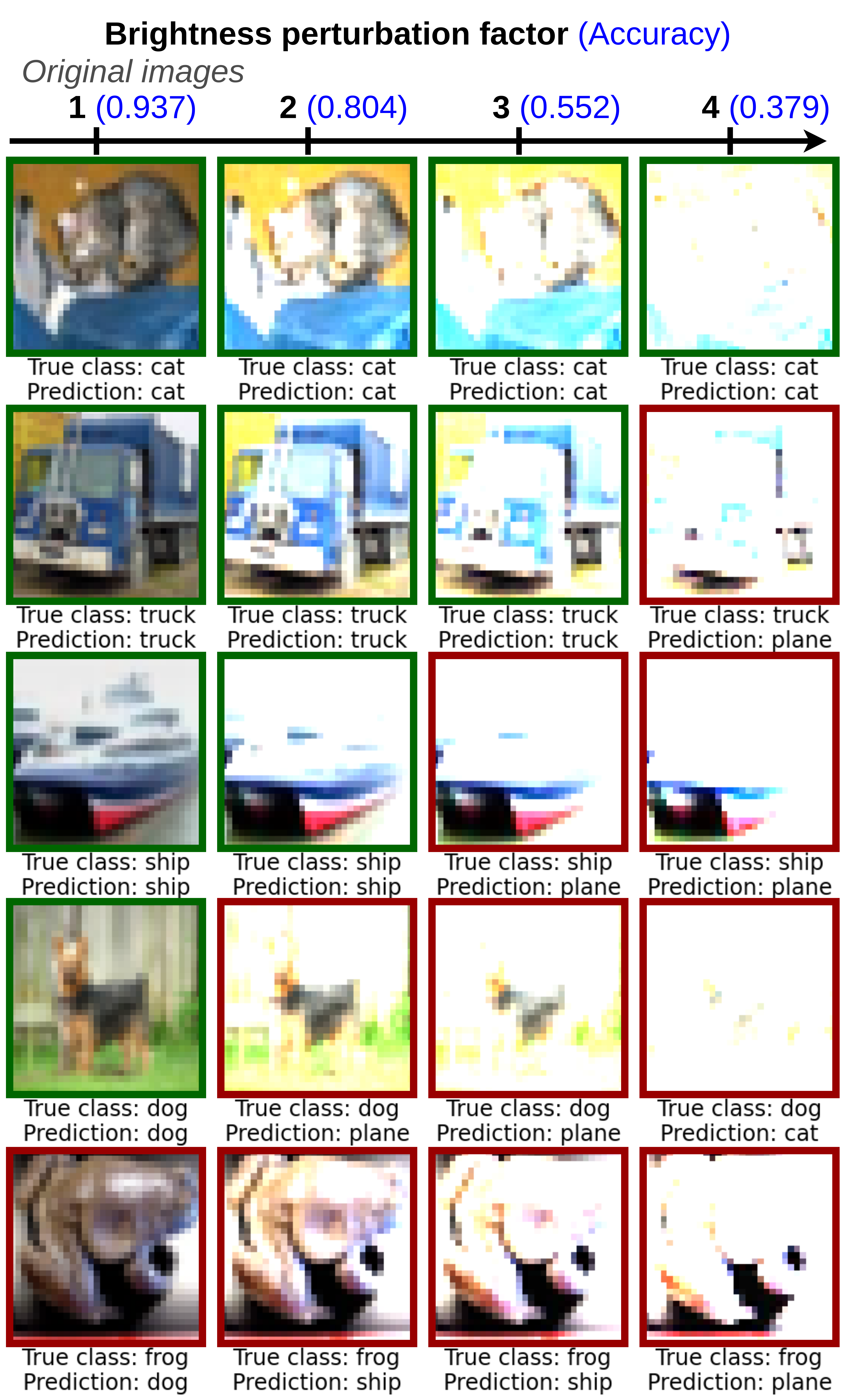}
    \caption{\textbf{What is Out-Of-Distribution?} Images from CIFAR10 test set with predictions by a ResNet-34 model. The columns represent increasing brightness perturbation levels.}
    \label{fig:brightness}
\end{figure}

\subsection{The definition of OOD is ambiguous}

\begin{figure*}[t]
    \centering
    \includegraphics[width=0.8\textwidth]{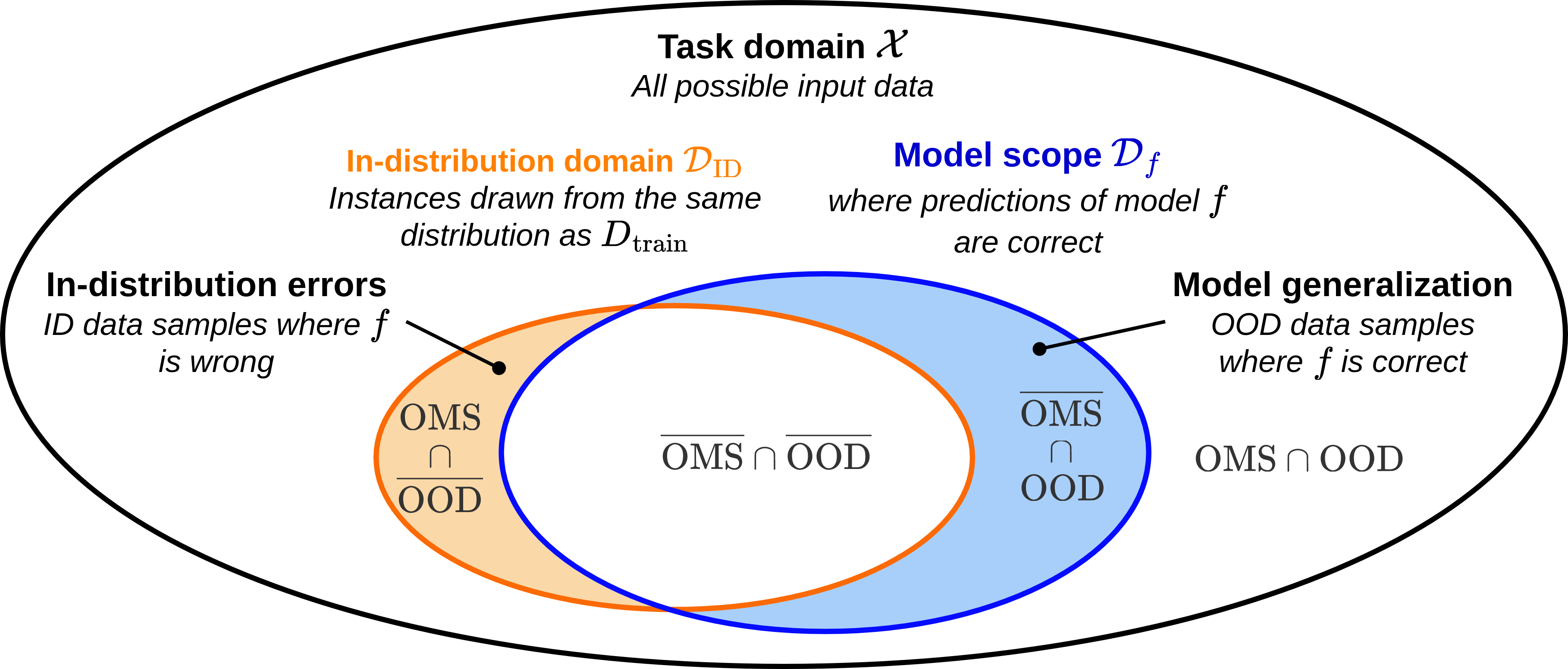}
    \caption{\textbf{Differences between OOD and OMS detection}. Schematic representation of the in-distribution domain $\mathcal{D}_{\text{ID}}$ for a task, as well as the model scope $\mathcal{D}_{f}$ of a model $f$ for this task.}
    \label{fig:ood_oms}
\end{figure*}

The first issue with OOD detection is that $\mathcal{D}_{\text{ID}}$ is not well-defined. Figure~\ref{fig:brightness} shows examples of brightness covariate shifts to illustrate this problem. Here, it is not clear how to set a threshold on the perturbation factor to define OODness. At a perturbation level of 2, the accuracy of $f$ has dropped about $13\%$, but it can still produce above $80\%$ of correct predictions. Hence, should the second column be considered OOD? This limitation also applies to adversarial examples, as different attacks lead to different levels of performance drop. For example, if we can generate a few additional misclassifications by changing random pixels, is it an adversarial attack? Regarding novelty detection, the problem is better defined: an image is OOD if its label is not among the output classes handled by $f$ (e.g., an image of a dinosaur is clearly OOD for CIFAR10). However, it is not easy to decide whether a highly corrupted image should hold its label. In figure~\ref{fig:brightness}, is the last image of the fourth row still a dog?

There is no simple answer to the questions raised in the previous paragraph. As a matter of fact, \citet{balestriero2021learning} showed that for high-dimensional data spaces, such as images, interpolation almost never happens. Even test data, which are traditionally considered ID, are not in the convex hull of $D_{\text{train}}$ and require extrapolation capabilities of $f$. In other words, the only data samples that are unambiguously ID are the training data themselves. This way, building an OOD evaluation dataset requires making subjective choices regarding the definition of OODness. 

Recently, \citet{wang2022vim} attempted to build a less ambiguous OOD benchmark dataset by asking two independent human annotators to label OOD images for ImageNet. The resulting OOD set contains all images labeled as OOD by both annotators. Unfortunately, this approach does not really solve the problem, as difficult images for humans are not necessarily what the DNN will struggle to process. In addition, even if this dataset is actually difficult for $f$, how can we know that it is not too simple for $m_f$? What about images that are less clearly OOD?

A less ambiguous way to discuss the concept of OOD is to use ideas from the field of DNN calibration~\cite{guo2017calibration}: a set of images is OOD if the accuracy of $f$ falls below a fixed user-defined threshold on this set. This definition is still threshold dependent, but it describes OODness unambiguously. However, under this definition, OODness depends on $f$ and is a property of a set, not defined at the individual sample level. This definition could be used to detect a performance drop over time by evaluating the OOD rate in consecutive data, but it is a different problem than the one studied in most OOD detection works.

\subsection{OOD does not always represent OMS}

Even if we had access to an unambiguous boundary of $\mathcal{D}_{\text{ID}}$, OOD and OMS detection would still remain distinct problems. In the schematic representation of figure~\ref{fig:ood_oms}, we can see that $\mathcal{D}_{\text{ID}}$ and $\mathcal{D}_{f}$ can differ in two different ways. 

\paragraph{Model generalization} The first difference, represented by the light blue region, is when input data are OOD but correctly classified by $f$ (e.g., the top-right image of figure~\ref{fig:brightness}). The limits of the ID domain are often defined by human programmers, independently of the generalization capabilities of $f$. For such cases, a good OOD monitor will reject perfectly valid inputs (false positives), thus decreasing the availability\footnote{In the dependability literature, the availability of a system represents its readiness to deliver a correct service.} of the model $f$. For the special case of novelty detection, this region does not exist because the model cannot be correct on data representing novel classes.

\paragraph{In-distribution errors} The second difference between OOD and OMS lies in the orange region in figure~\ref{fig:ood_oms}. In-distribution errors are data samples that are similar to training data, but that are misclassified by $f$ (e.g., the bottom-left image of figure~\ref{fig:brightness}). Such cases can have catastrophic outcomes for safety-critical applications, since if an OOD monitor accepts wrong predictions as long as they are similar to $D_{\text{train}}$, it also let hazardous predictions through the system (false negatives) and decreases its safety. This problem is an issue even when considering novelty detection. 

Furthermore, it is counter-productive to expect a monitor to accept misclassified training data. If a data sample is misclassified, its representation by the DNN is likely not similar to other representatives of its class, and forcing the monitor to say otherwise does not make sense. 

\subsection{Switching to OMS}

The above discussion showed that it is hard to come up with an objective definition of what OOD means, and that building good OOD monitors might not be sufficient to ensure the safety of a system using a DNN. The good news is that such a definition is not necessary, as we can simply use OMS as a generic setting for studying DNN runtime monitoring. The OMS paradigm is defined unambiguously and actually corresponds to what we want to achieve, i.e., reject wrong predictions of $f$. In practice, it is common to define a proxy task (here OOD detection) to address a more complex generic task of interest (OMS detection). However, when the actual performance at the true task is easy to compute, we should use it to evaluate the performance of new techniques. In our case, computing OMS detection performance is easy as we have access to labeled datasets and we can compute the predictions of $f$. For these reasons, we argue that DNN runtime monitors should be evaluated with respect to the OMS setting, instead of the ambiguous OOD setting.


\section{Experimental validation} \label{sec:differences_expe}
So far, we have argued that 
if the ultimate goal is to increase DNN safety, we should always evaluate runtime monitors under the OMS setting. However, many recent works have reported the performance of their approaches under the OOD setting (Section~\ref{sec:related_ood}). In this section, we conduct experiments on common OOD datasets from the literature, to see whether analyzing OOD results can lead to misleading conclusions about the safety of the system.


\subsection{Experiments description} \label{sec:differences_expe_description}
Our experiments consist of 27 OOD datasets, 2 DNN architectures (54 OOD scenarios), and 6 monitoring approaches.

We use three popular image datasets as ID: CIFAR10, CIFAR100, and SVHN \cite{krizhevsky2009cifar, netzer2011svhn}. For each ID dataset, the train split is used to fit the monitors, while the test split serves as the ID set for evaluation. Each ID set is combined with 9 distinct OOD sets:
\begin{itemize}
    \item Three novelty tasks -- For CIFAR10 we use CIFAR100, SVHN, and LSUN~\cite{yu2015lsun} to represent novel data. For CIFAR100 we use CIFAR10, SVHN, and LSUN. For SVHN we use CIFAR10, LSUN, and TinyImageNet (subset of ImageNet~\cite{deng2009imagenet}). With these choices, ID and OOD classes never overlap.
    \item Three covariate shift tasks -- For each ID dataset, we apply three image transformations from the AugLy library~\cite{papakipos2022augly}: Brightness (factor=3), Blur (radius=2), and Pixelization (ratio=0.5).
    \item Three adversarial attacks -- For each ID dataset, we apply three adversarial attacks from Torchattacks~\cite{kim2020torchattacks}, with default parameters: FGSM, DeepFool, and PGD.
\end{itemize}
For each of these 27 OOD scenarios, we experiment with two DNN architectures: DenseNet and ResNet. We use the pre-trained models from \citet{lee2018mahalanobis}.

For monitoring, we test two feature-based approaches: Mahalanobis (\textbf{Maha})~\cite{lee2018mahalanobis} and Outside-the-Box (\textbf{OtB})~\cite{henzinger2020outside}. We use the last layer before classification as their data representation. We also use four logit-based approaches: Max Softmax Probability (\textbf{MSP}) \cite{hendrycks17baseline}, Energy (\textbf{Ene})~\cite{liu2020energy}, and ReAct combined with both MSP (\textbf{R-MSP}) and Energy (\textbf{R-Ene})~\cite{sun2021react}.
Except for OtB, the techniques used in our experiments require selecting a rejection threshold on the monitoring scores. This is a complex question that is not studied here. Instead, we consider the optimal F1 threshold, as described by \citet{guerin2020robust}.

The complete code to reproduce our experiments can be found at \url{https://github.com/jorisguerin/neural-network-monitoring-benchmark}

\begin{figure}[t]
    \centering
    \includegraphics[width=0.49\textwidth]{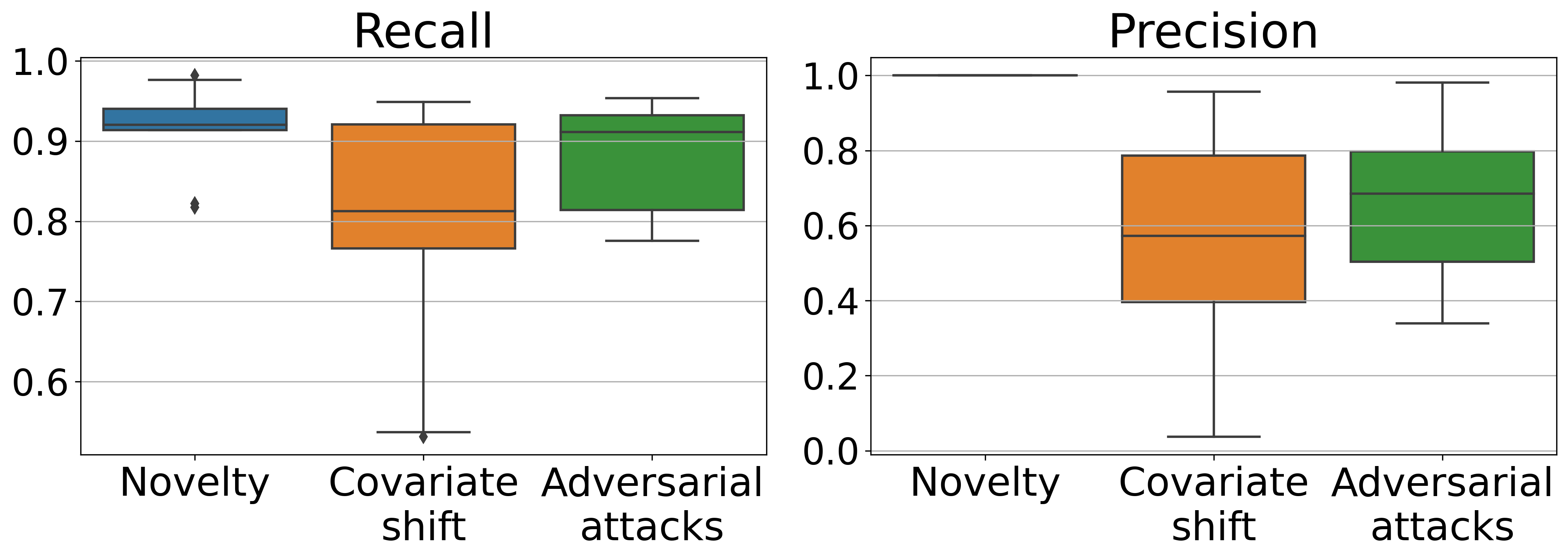}
    \caption{\textbf{OMS results for optimal OOD monitors}. Distribution of the OMS recall and precision obtained by the optimal OOD detector $m^*$ across the 27 OOD datasets and two neural networks tested in our experiments.}
    \label{fig:optOOD}
\end{figure}

\subsection{OOD can give a false sense of safety} \label{sec:differences_expe_optimal}
Here, we want to answer the following question: if we manage to develop a perfect OOD detector $m^*$, can we guarantee that ($f$, $m^*$) is safe to use in critical applications? 

Let's suppose that we are able to build $m^*$, which rejects all OOD samples and accepts all ID samples (precision and recall of 1 at OOD detection). In our experiments, we can easily simulate the predictions of $m^*$ using the known binary ID/OOD labels. Then, we want to know how well $m^*$ performs the task of detecting OMS data samples. An OMS recall below one means that there exist ID data points for which $f$ is erroneous. It represents a threat to the safety of the system. An OMS precision below one means that there exist OOD data points for which $f$ is correct. It represents a decrease in the availability of the system.

We conduct this experiment for both DNN architectures, across the 27 OOD datasets. The distribution of the OMS performances obtained for the different OOD types is reported in Figure~\ref{fig:optOOD}. We can see that the OMS recall is not close to 1 for most of the experiments conducted. Even for novelty, the median OMS recall of the 18 experiments is below 0.93. In other words, in most cases, more than $7\%$ of the errors of $f$ are not detected by the perfect OOD monitor $m^*$, which can lead to catastrophic outcomes for safety-critical applications. This effect is even more pronounced for covariate shifts and adversarial attacks. 

\begin{figure*}[t]
    \centering
\begin{subfigure}[b]{0.49\textwidth}
    \centering
    \includegraphics[width=\textwidth]{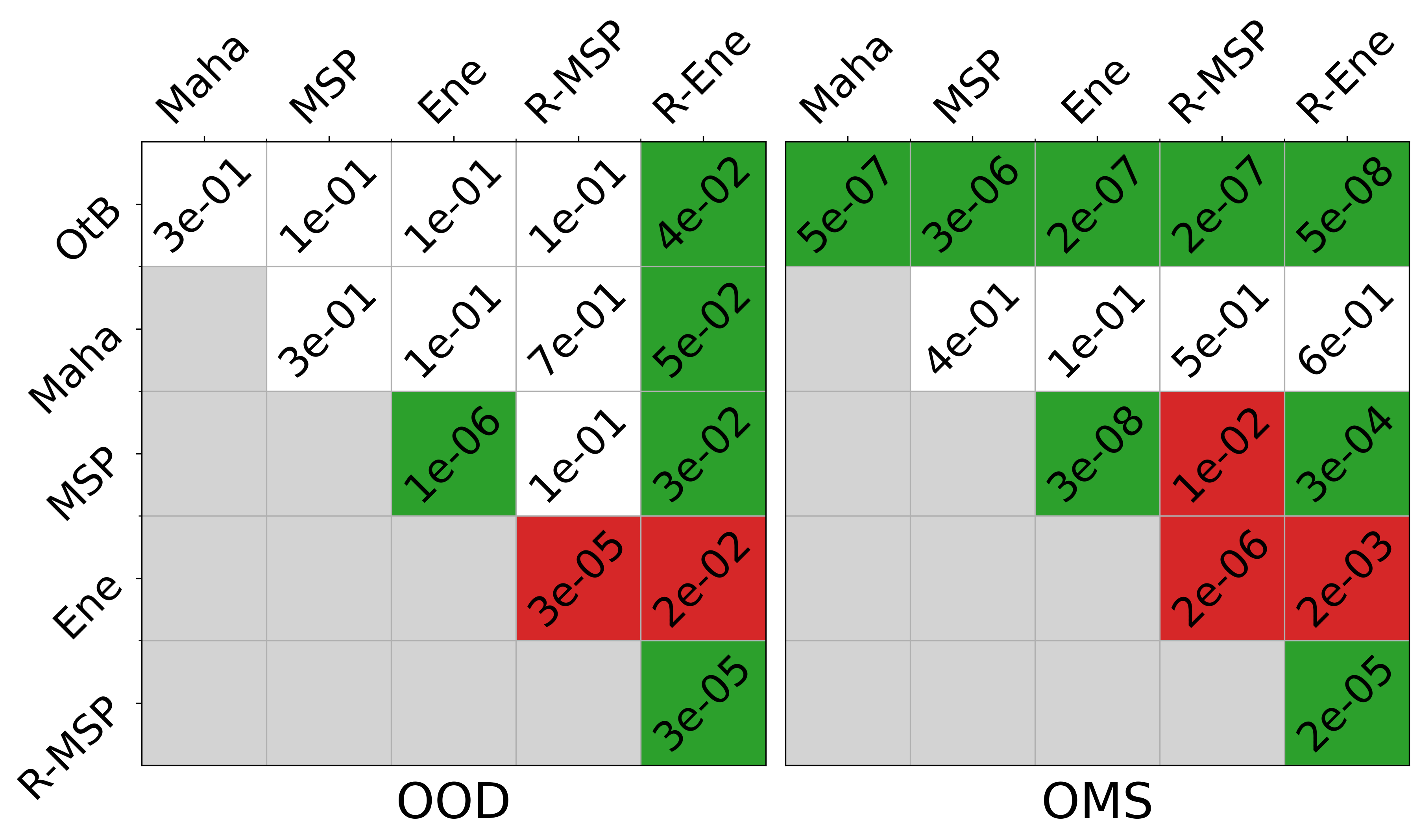}
    \caption{Recall}
    \label{fig:literature_oodOms_recall}
\end{subfigure}
~
\begin{subfigure}[b]{0.49\textwidth}
    \centering
    \includegraphics[width=\textwidth]{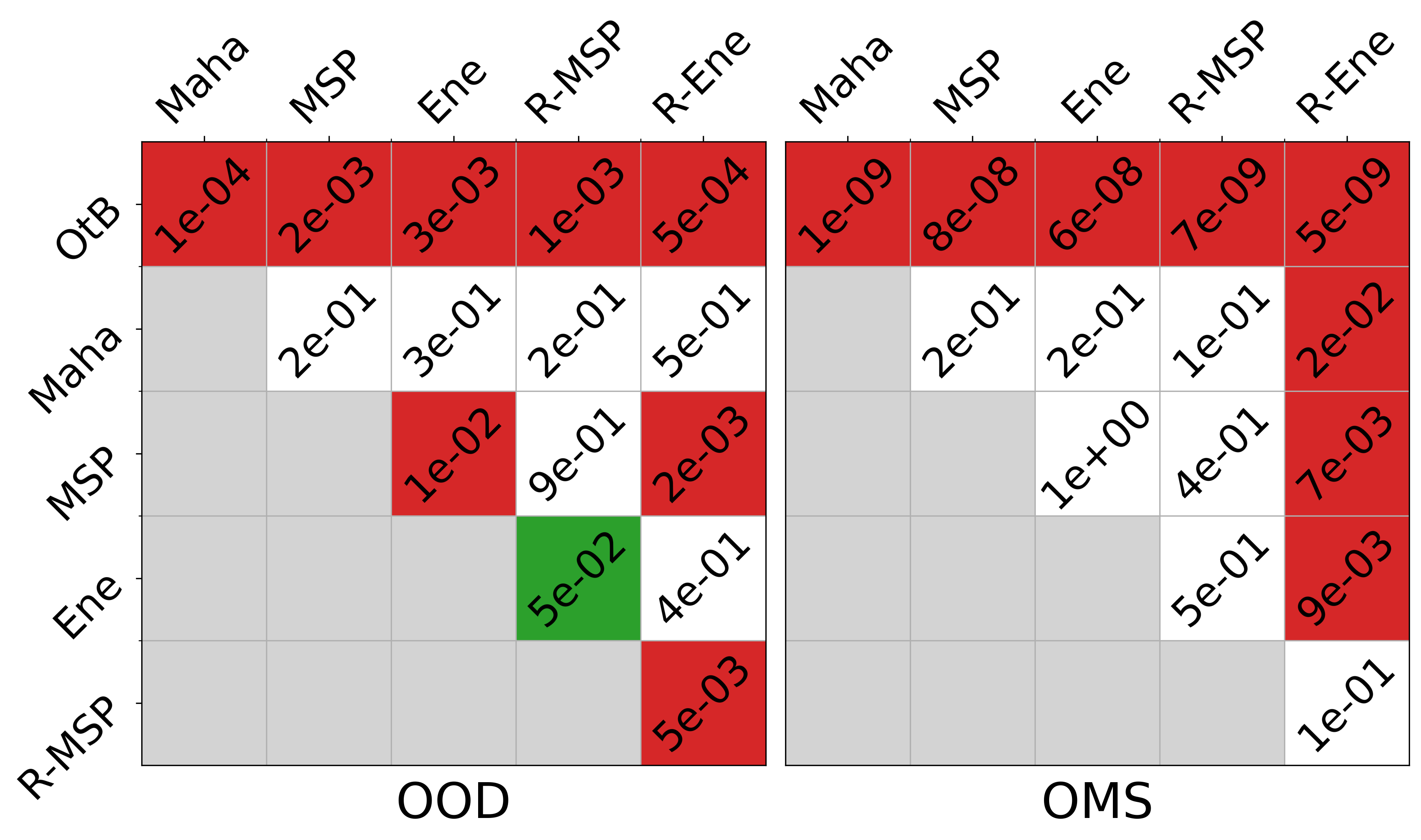}
    \caption{Precision}
    \label{fig:literature_oodOms_precision}
\end{subfigure}
\caption{\textbf{Pairwise comparison of monitors}. Popular monitors are compared across 54 OOD scenarios using a Wilcoxon test. The obtained p-values are reported. Green cells indicate that the approach corresponding to the row is better than the column, red cells indicate that the column is better than the row, and white cells indicate no statistically significant difference ($p>0.05$).}
\label{fig:literature_oodOms}
\end{figure*}

These results show that, even if perfect OOD detectors are developed, it will not allow us to guarantee the safety of critical systems using ML. Indeed, an OOD recall of 1 does not guarantee the absence of prediction errors. Even worse, reporting very good OOD results can be detrimental as it gives a false sense that the ($f$, $m^*$) system is safe.

Regarding OMS precision, we first note that it is always 1 for novelty datasets. This is because $f$ cannot be correct for novel samples, by definition. For other OOD types, we acknowledge an important drop in precision, i.e., an important proportion of correct predictions of $f$ on OOD data.

    

\subsection{OOD can be misleading for comparison} \label{sec:differences_expe_monitors}

As explained in section~\ref{sec:related_ood}, monitors are often compared using the OOD setting. Here, we want to verify whether comparing monitors at the OOD detection task provides relevant insights into OMS detection performance. 

To do this, we compare six monitors from the literature across 27 OOD datasets and two DNNs architectures (section~\ref{sec:differences_expe_description}). For each pair of monitors, we conduct a Wilcoxon signed-rank test across the 54 OOD scenarios to determine whether one is better than the other. The Wilcoxon test is a non-parametric statistical test that can be used to compare pairs of classifiers across multiple datasets~\cite{demvsar2006statistical}. The results obtained are reported in figure~\ref{fig:literature_oodOms}.

From these experiments, we can see that the comparison matrices obtained for OOD and OMS look different. For example, OtB is clearly the most conservative approach (highest recall) when looking at the OMS results, but not when considering OOD. On another note, the benefits of using ReAct for monitoring are best seen in the OMS setting (MSP vs. R-MSP, Ene vs. R-Ene).

From these results, it appears that conducting OOD experiments is not a reliable way to compare monitoring approaches regarding their ability to identify errors of $f$.

    


\section{Removing training outliers to build better runtime monitors} \label{sec:recommendations}

As explained earlier (Figure~\ref{fig:RM_def}), runtime monitors are often fitted using the DNN training dataset to represent ``what is safe''. The common practice in the literature is to use all available $D_{\text{train}}$ samples to fit the monitor. When considering OOD, the objective is to build a binary classifier that can reject ``data samples from outside of the training distribution''  (dissimilar to $D_{\text{train}}$). Hence, in this context, it makes sense to fit the monitor using all available training data samples. 

In practice, DNNs rarely perform perfectly, even on the exact data used for training. For example, the ResNet-34 used in this work misclassifies $127$ of the $50,000$ training data samples (training accuracy of $99.746\%$). When considering the OMS setting, the objective is to build a classifier to identify data samples that are misclassified by $f$. Hence, it is conceptually wrong to use the misclassified training samples to represent the normal class when fitting a one-class classifier for OMS detection. In this section, we propose a simple trick to build better DNN monitors, consisting in removing erroneous training samples to fit the monitor. 

\begin{figure*}[t]
    \centering
    \includegraphics[width=0.9\textwidth]{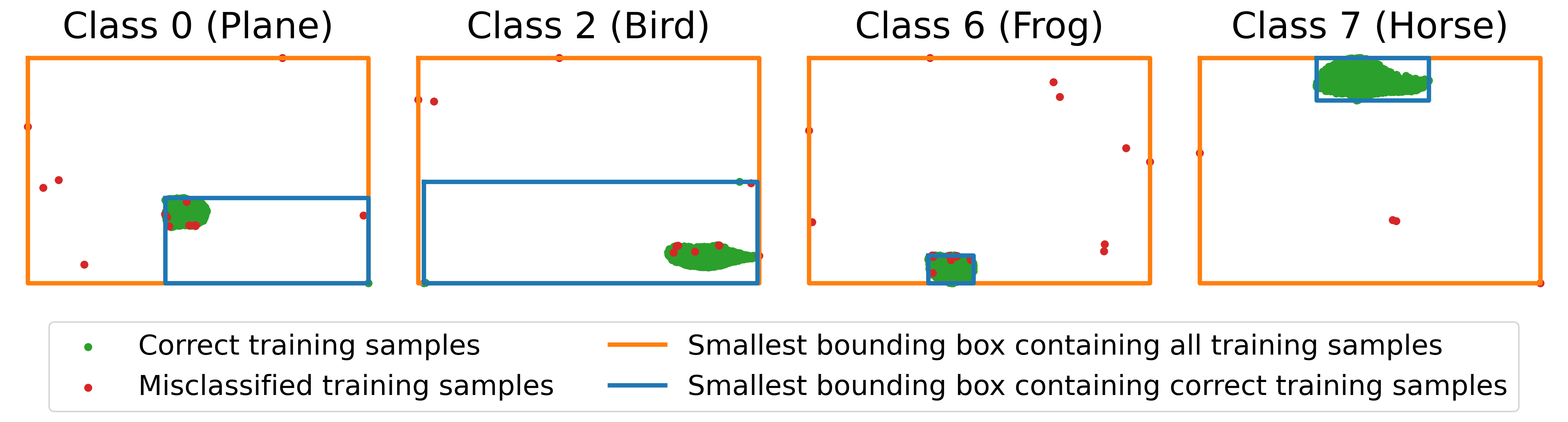}
    \caption{\textbf{Should we use misclassified training samples to fit a monitor?} Two-dimensional UMAP visualization of four selected classes from the CIFAR10 training split. Colors indicate whether they were correctly classified by the ResNet-34 model. We also show the minimal bounding boxes containing all training samples (orange) and all correctly classified training samples (blue). Misclassified training samples are often outliers, leading to much larger bounding boxes when they are included.}
    \label{fig:umap}
\end{figure*} 

To illustrate this idea, we apply a two-dimensional Uniform Manifold Approximation and Projection (UMAP) transformation~\cite{mcinnes2018umap} to the training split of CIFAR10. UMAP is an unsupervised embedding method that produces a lower-dimensional representation preserving the global distances. 
The 2D representation obtained allows us to investigate visually whether the misclassified training samples are outliers, which change the shape of the one-class classifier boundary. The embeddings of four CIFAR10 classes are shown in figure~\ref{fig:umap}. It shows that correctly classified points form compact clusters while misclassified points appear to be clear outliers. To further illustrate our point, we draw the minimal bounding box containing all training samples, and the one containing only correctly classified samples. These bounding boxes represent the one-class classification boundaries of an Outside-the-Box monitor without K-means pre-processing~\cite{henzinger2020outside}. The blue boxes, obtained by removing misclassified samples, are much tighter than the orange ones, and appear to better represent the model scope.

Figure~\ref{fig:umap} suggests that removing misclassified training data helps to fit better monitors. However, UMAP embeddings are not real data and one should not rely on intuition gained from such visualizations. Hence, we conduct more formal experiments and compare the performance of a simple OtB monitor (without K-means) when trained with and without misclassified training samples. The mean and standard deviation (St. dev.) of precision and recall are computed across all 54 OOD scenarios (section~\ref{sec:differences_expe}) and reported in table~\ref{tab:results_trick}. We also conduct a Wilcoxon test and report the p-values obtained. These results demonstrate that removing misclassified training samples can help to increase the recall (clearly statistically significant) without impacting the precision too much. In other words, the monitor can detect more errors of $f$ without discarding additional valid predictions.


\begin{table}[t]
    \centering
    \begin{tabular}{ccc}
    \toprule
        & \textbf{Recall} & \textbf{Precision}\\
        & Mean (St. dev.) & Mean (St. dev.) \\
     \toprule
    \textbf{All training data} & 0.75 (0.19) & 0.68 (0.18)\\
    \textbf{Only correct data} & 0.77 (0.18) & 0.68 (0.18)\\
    \midrule
    \textbf{Wilcoxon test} & Better & Worst \\
    p-value & 2.6e-9 & 1.6e-3 \\
    \bottomrule
    \end{tabular}
    \caption{OMS results for Outside-the-box}
    \label{tab:results_trick}
\end{table}

Two previous works already had the intuition that removing misclassified training samples could reduce the false negative rate \cite{henzinger2020outside, cheng2019runtime}. This section explains this intuition and shows experimental evidence of the benefits of this simple trick in the context of OMS detection. 

In these experiments, we only used OtB because it is a threshold-free approach. As threshold selection strategies were not discussed in this work, other monitors were not considered here. However, when trying to select an optimal monitoring threshold, we believe that it would also be valuable to not consider misclassified samples as normal data. 


\section{Conclusion}
If we want to use DNN in safety-critical applications, it is paramount to build efficient runtime monitors, which can detect and remove DNN errors from the system before they can have catastrophic consequences. As of today, many research efforts are directed toward solving the problem of out-of-distribution detection, which consists in identifying data samples that were drawn from a different distribution than the DNN training set. In this paper, we discuss the limitations of this setting to enable the safe usage of DNN in critical systems. First, the concept of OOD is not well-defined, which makes it difficult to compare different OOD detection approaches. Second, even a perfect OOD detector can have the undesirable property of discarding valid predictions of the DNN, and even worse, accepting wrong predictions. Extensive experiments conducted on popular OOD detection datasets confirm that these phenomena occur in practical scenarios. Furthermore, we show that the OOD setting cannot be used to compare monitoring approaches accurately, as the best monitor for OOD is not always the best one to detect DNN errors. Finally, we demonstrate experimentally that it is a good practice to remove training data points that the DNN cannot classify correctly before fitting the monitor.

On the bright side, adapting current OOD research to these findings will not require drastic changes. The out-of-model-scope (OMS) setting, defined above and discussed in several previous works, is well-defined and perfectly aligned with the DNN monitoring objectives. In addition, most approaches developed for OOD detection can be used for OMS without any modification, and OMS results can be computed straightforwardly for most OOD datasets used in the literature. Hence, the take-home message from this paper is that instead of evaluating new approaches on their ability to detect samples from other data sources, the OOD detection research community should focus on the ability to detect samples leading to erroneous DNN predictions. We also believe that it is a good idea to include OOD samples in OMS evaluation datasets, which is rarely done in OMS papers. It is worth mentioning that we recently introduced a unified evaluation formalism for runtime monitors, considering the entire system in which a DNN is included~\cite{guerin2022unifying}.


\section*{Acknowledgments}
This research has benefited from the AI Interdisciplinary Institute ANITI. ANITI is funded by the French ”Investing for the Future – PIA3” program under the Grant Agreement No ANR-19-PI3A-0004.

This research has also received funding from the European Union’s Horizon 2020 research and innovation program under the Marie Skłodowska-Curie grant agreement No 812.788 (MSCA-ETN SAS). This publication reflects only the authors’ view, exempting the European Union from any liability. Project website: http://etn-sas.eu/.


\begin{thebibliography}{35}
\providecommand{\natexlab}[1]{#1}

\bibitem[{Balestriero, Pesenti, and LeCun(2021)}]{balestriero2021learning}
Balestriero, R.; Pesenti, J.; and LeCun, Y. 2021.
\newblock Learning in high dimension always amounts to extrapolation.
\newblock \emph{arXiv preprint arXiv:2110.09485}.

\bibitem[{Cai and Koutsoukos(2020)}]{cai2020real}
Cai, F.; and Koutsoukos, X. 2020.
\newblock Real-time out-of-distribution detection in learning-enabled
  cyber-physical systems.
\newblock In \emph{2020 ACM/IEEE 11th International Conference on
  Cyber-Physical Systems (ICCPS)}, 174--183. IEEE.

\bibitem[{Chen, Yoon, and Shao(2020)}]{chen2020task}
Chen, V.; Yoon, M.-K.; and Shao, Z. 2020.
\newblock Task-Aware Novelty Detection for Visual-based Deep Learning in
  Autonomous Systems.
\newblock In \emph{2020 IEEE International Conference on Robotics and
  Automation (ICRA)}, 11060--11066. IEEE.

\bibitem[{Cheng, N{\"u}hrenberg, and Yasuoka(2019)}]{cheng2019runtime}
Cheng, C.-H.; N{\"u}hrenberg, G.; and Yasuoka, H. 2019.
\newblock Runtime monitoring neuron activation patterns.
\newblock In \emph{2019 Design, Automation \& Test in Europe Conference \&
  Exhibition (DATE), Florence, Italy}, 300--303. IEEE.

\bibitem[{Dem{\v{s}}ar(2006)}]{demvsar2006statistical}
Dem{\v{s}}ar, J. 2006.
\newblock Statistical comparisons of classifiers over multiple data sets.
\newblock \emph{The Journal of Machine learning research}, 7: 1--30.

\bibitem[{Deng et~al.(2009)Deng, Dong, Socher, Li, Li, and
  Fei-Fei}]{deng2009imagenet}
Deng, J.; Dong, W.; Socher, R.; Li, L.-J.; Li, K.; and Fei-Fei, L. 2009.
\newblock Imagenet: A large-scale hierarchical image database.
\newblock In \emph{2009 IEEE conference on computer vision and pattern
  recognition}, 248--255. Ieee.

\bibitem[{Ferreira et~al.(2021)Ferreira, Arlat, Guiochet, and
  Waeselynck}]{ferreira2021benchmarking}
Ferreira, R.~S.; Arlat, J.; Guiochet, J.; and Waeselynck, H. 2021.
\newblock Benchmarking Safety Monitors for Image Classifiers with Machine
  Learning.
\newblock In \emph{2021 IEEE 26th Pacific Rim International Symposium on
  Dependable Computing (PRDC)}, 7--16. IEEE.

\bibitem[{Granese et~al.(2021)Granese, Romanelli, Gorla, Palamidessi, and
  Piantanida}]{granese2021doctor}
Granese, F.; Romanelli, M.; Gorla, D.; Palamidessi, C.; and Piantanida, P.
  2021.
\newblock DOCTOR: A Simple Method for Detecting Misclassification Errors.
\newblock \emph{Advances in Neural Information Processing Systems}, 34:
  5669--5681.

\bibitem[{Gu{\'e}rin, de~Paula~Canuto, and Goncalves(2020)}]{guerin2020robust}
Gu{\'e}rin, J.; de~Paula~Canuto, A.~M.; and Goncalves, L. M.~G. 2020.
\newblock Robust Detection of Objects under Periodic Motion with Gaussian
  Process Filtering.
\newblock In \emph{2020 19th IEEE International Conference on Machine Learning
  and Applications (ICMLA)}, 685--692. IEEE.

\bibitem[{Guerin, Delmas, and Guiochet(2022)}]{guerin2022evaluation}
Guerin, J.; Delmas, K.; and Guiochet, J. 2022.
\newblock Evaluation of Runtime Monitoring for UAV Emergency Landing.
\newblock In \emph{International Conference on Robotics and Automation (ICRA)},
  To appear. IEEE.

\bibitem[{Guerin et~al.(2022)Guerin, Ferreira, Delmas, and
  Guiochet}]{guerin2022unifying}
Guerin, J.; Ferreira, R.~S.; Delmas, K.; and Guiochet, J. 2022.
\newblock Unifying Evaluation of Machine Learning Safety Monitors.
\newblock In \emph{the 33rd IEEE International Symposium on Software
  Reliability Engineering (ISSRE 2022)}, To appear. IEEE.

\bibitem[{Guo et~al.(2017)Guo, Pleiss, Sun, and
  Weinberger}]{guo2017calibration}
Guo, C.; Pleiss, G.; Sun, Y.; and Weinberger, K.~Q. 2017.
\newblock On calibration of modern neural networks.
\newblock In \emph{International conference on machine learning}, 1321--1330.
  PMLR.

\bibitem[{Haidegger(2019)}]{surgical_robotics}
Haidegger, T. 2019.
\newblock Autonomy for surgical robots: Concepts and paradigms.
\newblock \emph{IEEE Transactions on Medical Robotics and Bionics}, 1(2):
  65--76.

\bibitem[{Hendrycks and Dietterich(2019)}]{hendrycks2019benchmarking}
Hendrycks, D.; and Dietterich, T. 2019.
\newblock Benchmarking Neural Network Robustness to Common Corruptions and
  Perturbations.
\newblock In \emph{International Conference on Learning Representations}.

\bibitem[{Hendrycks and Gimpel(2017)}]{hendrycks17baseline}
Hendrycks, D.; and Gimpel, K. 2017.
\newblock A Baseline for Detecting Misclassified and Out-of-Distribution
  Examples in Neural Networks.
\newblock \emph{Proceedings of International Conference on Learning
  Representations}.

\bibitem[{Henzinger, Lukina, and Schilling(2020)}]{henzinger2020outside}
Henzinger, T.~A.; Lukina, A.; and Schilling, C. 2020.
\newblock Outside the Box: Abstraction-Based Monitoring of Neural Networks.
\newblock In \emph{24th European Conference on Artificial Intelligence-ECAI
  2020}, 2433--2440.

\bibitem[{Kang et~al.(2018)Kang, Raghavan, Bailis, and Zaharia}]{kang2018model}
Kang, D.; Raghavan, D.; Bailis, P.; and Zaharia, M. 2018.
\newblock Model assertions for debugging machine learning.
\newblock In \emph{NeurIPS MLSys Workshop}, volume~3, 10.

\bibitem[{Kantaros et~al.(2021)Kantaros, Carpenter, Sridhar, Yang, Lee, and
  Weimer}]{kantaros2021real}
Kantaros, Y.; Carpenter, T.; Sridhar, K.; Yang, Y.; Lee, I.; and Weimer, J.
  2021.
\newblock Real-time detectors for digital and physical adversarial inputs to
  perception systems.
\newblock In \emph{Proceedings of the ACM/IEEE 12th International Conference on
  Cyber-Physical Systems}, 67--76.

\bibitem[{Khan and Madden(2014)}]{khan2014occ_review}
Khan, S.~S.; and Madden, M.~G. 2014.
\newblock One-class classification: taxonomy of study and review of techniques.
\newblock \emph{The Knowledge Engineering Review}, 29(3): 345--374.

\bibitem[{Kim(2020)}]{kim2020torchattacks}
Kim, H. 2020.
\newblock Torchattacks: A pytorch repository for adversarial attacks.
\newblock \emph{arXiv preprint arXiv:2010.01950}.

\bibitem[{Krizhevsky, Hinton et~al.(2009)}]{krizhevsky2009cifar}
Krizhevsky, A.; Hinton, G.; et~al. 2009.
\newblock Learning multiple layers of features from tiny images.

\bibitem[{Lee et~al.(2018)Lee, Lee, Lee, and Shin}]{lee2018mahalanobis}
Lee, K.; Lee, K.; Lee, H.; and Shin, J. 2018.
\newblock A simple unified framework for detecting out-of-distribution samples
  and adversarial attacks.
\newblock \emph{Advances in neural information processing systems}, 31.

\bibitem[{Liang, Li, and Srikant(2018)}]{liang2018odin}
Liang, S.; Li, Y.; and Srikant, R. 2018.
\newblock Enhancing The Reliability of Out-of-distribution Image Detection in
  Neural Networks.
\newblock In \emph{International Conference on Learning Representations}.

\bibitem[{Liu et~al.(2020)Liu, Wang, Owens, and Li}]{liu2020energy}
Liu, W.; Wang, X.; Owens, J.; and Li, Y. 2020.
\newblock Energy-based out-of-distribution detection.
\newblock \emph{Advances in Neural Information Processing Systems}, 33:
  21464--21475.

\bibitem[{McInnes, Healy, and Melville(2018)}]{mcinnes2018umap}
McInnes, L.; Healy, J.; and Melville, J. 2018.
\newblock Umap: Uniform manifold approximation and projection for dimension
  reduction.
\newblock \emph{arXiv preprint arXiv:1802.03426}.

\bibitem[{Netzer et~al.(2011)Netzer, Wang, Coates, Bissacco, Wu, and
  Ng}]{netzer2011svhn}
Netzer, Y.; Wang, T.; Coates, A.; Bissacco, A.; Wu, B.; and Ng, A.~Y. 2011.
\newblock Reading Digits in Natural Images with Unsupervised Feature Learning.
\newblock In \emph{NIPS Workshop on Deep Learning and Unsupervised Feature
  Learning 2011}.

\bibitem[{Papakipos and Bitton(2022)}]{papakipos2022augly}
Papakipos, Z.; and Bitton, J. 2022.
\newblock AugLy: Data Augmentations for Robustness.
\newblock arXiv:2201.06494.

\bibitem[{Schorn and Gauerhof(2020)}]{schorn2020facer}
Schorn, C.; and Gauerhof, L. 2020.
\newblock FACER: A universal framework for detecting anomalous operation of
  deep neural networks.
\newblock In \emph{2020 IEEE 23rd International Conference on Intelligent
  Transportation Systems (ITSC)}, 1--6. IEEE.

\bibitem[{Stocco et~al.(2020)Stocco, Weiss, Calzana, and
  Tonella}]{stocco2020misbehaviour}
Stocco, A.; Weiss, M.; Calzana, M.; and Tonella, P. 2020.
\newblock Misbehaviour prediction for autonomous driving systems.
\newblock In \emph{Proceedings of the ACM/IEEE 42nd International Conference on
  Software Engineering}, 359--371.

\bibitem[{Sun, Guo, and Li(2021)}]{sun2021react}
Sun, Y.; Guo, C.; and Li, Y. 2021.
\newblock React: Out-of-distribution detection with rectified activations.
\newblock \emph{Advances in Neural Information Processing Systems}, 34:
  144--157.

\bibitem[{Wang et~al.(2022)Wang, Li, Feng, and Zhang}]{wang2022vim}
Wang, H.; Li, Z.; Feng, L.; and Zhang, W. 2022.
\newblock ViM: Out-Of-Distribution with Virtual-logit Matching.
\newblock In \emph{Proceedings of the IEEE/CVF Conference on Computer Vision
  and Pattern Recognition}, 4921--4930.

\bibitem[{Wang et~al.(2020)Wang, Xu, Xu, Ma, and Lu}]{wang2020dissector}
Wang, H.; Xu, J.; Xu, C.; Ma, X.; and Lu, J. 2020.
\newblock Dissector: Input validation for deep learning applications by
  crossing-layer dissection.
\newblock In \emph{2020 IEEE/ACM 42nd International Conference on Software
  Engineering (ICSE)}, 727--738. IEEE.

\bibitem[{Wang et~al.(2019)Wang, Dong, Sun, Wang, and
  Zhang}]{wang2019adversarial}
Wang, J.; Dong, G.; Sun, J.; Wang, X.; and Zhang, P. 2019.
\newblock Adversarial sample detection for deep neural network through model
  mutation testing.
\newblock In \emph{2019 IEEE/ACM 41st International Conference on Software
  Engineering (ICSE)}, 1245--1256. IEEE.

\bibitem[{Yu et~al.(2015)Yu, Seff, Zhang, Song, Funkhouser, and
  Xiao}]{yu2015lsun}
Yu, F.; Seff, A.; Zhang, Y.; Song, S.; Funkhouser, T.; and Xiao, J. 2015.
\newblock Lsun: Construction of a large-scale image dataset using deep learning
  with humans in the loop.
\newblock \emph{arXiv preprint arXiv:1506.03365}.

\bibitem[{Zhang et~al.(2018)Zhang, Zhang, Zhang, Liu, and
  Khurshid}]{zhang2018deeproad}
Zhang, M.; Zhang, Y.; Zhang, L.; Liu, C.; and Khurshid, S. 2018.
\newblock DeepRoad: GAN-based metamorphic testing and input validation
  framework for autonomous driving systems.
\newblock In \emph{2018 33rd IEEE/ACM International Conference on Automated
  Software Engineering (ASE)}, 132--142. IEEE.

\end{thebibliography}
\end{document}